          \documentclass[a4paper,12pt]{article}
\usepackage{amsmath}%
\usepackage{amsfonts}%
\usepackage{amssymb}%
\usepackage{graphicx}

\begin{document}

\title{Energy-Efficiency Path Planning for Quadrotor UAV Under Wind Conditions}
\author{Fouad Yacef, Nassim Rizoug and Laid Degaa  \thanks{Fouad Yacef is with Centre de D\'eveloppement des Technologies Avanc\'ees, Algiers, Algeria {\tt\small fyacef@cdta.dz}; Nassim Rizoug, Laid Degaa and Fouad Yacef are with Ecole Sup\'erieure des Techniques A\'eronautiques et de Construction Automobile, France {\tt\small nassim.rizoug@estaca.fr, laid.degaa@estaca.fr, fouad.yacef@estaca.fr}}}
\date{}
\maketitle

\begin{abstract}
Quadrotor unmanned aerial vehicles have a limited quantity of embedded energy. To preserve and guaranty the success of the UAV mission, we should manage energy consumption during the mission. In this study we introduce an optimization algorithm to minimize the consumed energy in quadrotor mission under windy conditions. The mechanical energy consumed by rotors of the flying vehicle is formulated with an efficiency function. Then, we formulate the energy minimization problem as an optimal control problem. The last problem is solved in order to calculate minimum energy for quadrotor simple mission under windy conditions. In simulation experiment, we compare the proposed method with an adaptive control approach.
\end{abstract}

\section{Introduction}
Unmanned Aerial Vehicles (UAVs), especially rotary wing unmanned aerial systems, used today in many application fields. An eminent case of use is to reach inaccessible areas autonomously, for example in reconnaissance missions \cite{watts2012unmanned}, the fields of search and rescue \cite{hildmann2019using} \cite{Sampedro2019}, inspection \cite{steich2016tree} and transportation of data and package \cite{tagliabue2017collaborative} are greatly interested in multi-copter utilization. One multi-copter has the ability to use embedded sensors, cameras or actuators to accomplish a wide range of tasks. Data collected from embedded sensors includes images and video, environmental measurements, and other types of data acquired from the exploration area. Due to the relatively simple mechanical design, the high maneuverability and hovering capabilities of multi-copters, these are especially useful for the mentioned fields of application. The configuration of multi-copters vary with the nature and constellation of wings, motors and propellers, which in return have an influence on the capabilities to accomplish the mission. All mentioned factors have a significant effect on the consumption of embedded energy for these aerial robots.

Energy efficiency is an important topic in the research field of aerial robots because this machines have a limited on-board energy. The small size and limited weight of battery for multi-copter provides a limited energy capacity, which results in a limitation of both overall flight time and flight distance. Possible solutions to limited flight autonomy are: (i) improving the efficiency of batteries or power sources, (ii) developing automated recharging methods, (iii) and improving the vehicles’ energy efficiency \cite{karydis2017energetics}.  Now, how can we improve energy efficiency?

To improve energy efficiency one can use algorithm-based optimization, or work on hardware-based optimization. A simple and easy way to reduce energy consumption is to minimize the weight of the vehicle by using for example airframes made of carbon fiber, and careful component selection of electronic devices (such as sensors, embedded computing) and batteries. Another way of this spirit is via mechanical redesign. Examples includes rotating quadrotor arms to a proper positions calculated based on the dynamics model of the quadrotor and the power-thrust curve of rotors \cite{xiong2019optimize}, or tilting the rotors about one or two orthogonal axes \cite{Ryll2015ANovel}\cite{Morbidi2018Energy}, or by the introduction of a more efficient “triangular” quadrotor \cite{Driessens2015heTriangular}. Algorithm-based optimization aims to develop novel path planning and control algorithms that consider energy consumption. Energy-efficiency algorithms reduce energy consumption and extend flight times through the design of minimum-energy trajectories tracked by aerial vehicles. 

 Algorithm-based optimization offers multiple opportunities for the improvement of efficiency of aerial machines, as it is easy to implement, economic to deploy, and can be used to complement mechanical designs, gaining insights for novel hardware platforms. Two principal approach can be used to achieve energy-efficiency algorithmic improvements; model-free and model-based approaches. A model-free approach \cite{Kreciglowa2019Energy}\cite{Tagliabue2019model}, allows taking into account effects that are difficult to model and less known, such as changes in performance due to aging of electronic components, or changes in the aerodynamic due to wind gust and payloads. A model-based approach \cite{vicencio2015energy}\cite{Ware2016Ananalysis}\cite{morbidi2016minimum} allows to fully exploiting the capabilities of the system, but relies on the ability to derive and identify an adequate model of the power consumption of a multi-copter. Such a model is usually focused on capturing the electrical power losses, or the aerodynamic power losses \cite{Bezzo2016Online}\cite{Franco2016coverage}\cite{yacef2017trajectory}\cite{lu2018tradeoff} of the robot.

Many solutions have been proposed recently contributing towards save energy, using model-based approach. In \cite{schacht2018path} a path planning algorithm that take into account the evolution of the battery performance is presented. the path planning algorithm is defined as an optimal control problem  with multi-objective optimization where the objective is to find a feasible trajectory between way-points whiles minimizing the energy consumed and the mission final time depending on the variation of the battery State of Health (SoH). Where in \cite{schacht2019mission} the effects of actuators fault occurring
during mission on energy consumption for multirotor UAV is analyzed. The impact of battery discharge,
State of Charge (SoC) and State of Health (SoH) during the mission execution is evaluated also, with
the actuators fault modeled as loss of effectiveness. The authors in \cite{Morbidi2018Energy} consider a non-conventional hexarotor whose propellers can be simultaneously tilted about two orthogonal axes. For a given tilt profile, the minimum-energy trajectory between two prescribed boundary states is explicitly determined by solving an optimal control problem with respect to the angular accelerations of the six brushless motors.

In this paper, we gave the power lose model with actuator and battery model of quadrotor UAV in the first place, then we introduced the energy optimal control problem, where the objective is to calculate minimum energy consumed by the quadrotor vehicle for simple mission under windy conditions. Moreover, we  generated energy-efficiency paths and rotor angular accelerations as control inputs by solving the optimal control problem. With the power lose model, we can compute values of energy consumption during the flight. We can finally evaluate the amount of energy saved during the mission by compared energy consumption for the proposed approach with an adaptive control approach.

The rest of this paper is organized as follows. Sect. II presents power lose model and brushless DC motor with battery model of a quadrotor UAV. In Sect. III, we discusses the mission and formulates the optimization problems, and in Sect. IV, numerical experiments is presented and its results are discussed. Finally, we summarizes conclusion and outline some promising future work Sect. V.  
\section{Power lose model}
There are two principal sources of power lose in quadrotor UAV mission. The first one is the on-board computer with sensors and the second one is the rotors. In this work we consider that the energy consumed by the on-board computer and embedded sensors is neglected compared to the one consumed by the four rotors.  
\subsection{Rotors and battery dynamic}
Quadrotors UAV are driven by brushless direct current (BLDC) motors, which can be modeled as electrical direct current DC motors. A simple model for DC motors can be represented by a circuit containing a resistor, inductor, and voltage generator in series \cite{yacef2017trajectory}.
\begin{equation} \label{eq:1}
\begin{cases}
I_r\dot{\omega}(t)=\tau(t)-\kappa\omega^2(t)\\
v(t)=Ri(t)+L\frac{\partial i(t)}{\partial t}+\frac{1}{k_v}\omega(t)
\end{cases}	
\end{equation}
where $R$ is the motor internal resistance, $L$ is the inductance, $\omega(t)$ is the rotor velocity, and $k_v$ is the voltage constant of the motor, expressed in $rad/s/volt$. The inertia, $I_r$ includes the motor and the propeller, the motor torque comes from the voltage generator, and the load friction torque results from the propeller drag $Q_f\big(\omega(t)\big)=\kappa\omega^2(t)$, $\kappa$ is the drag coefficient. Also, the motor torque $\tau(t)$ can be modeled as being proportional to the current $i(t)$ through the torque constant, $k_t$, expressed in $Nm/A$. 
\begin{equation} \label{eq:2}
\tau(t)=k_ti(t)
\end{equation}
Typically, the inductance of small, DC motors is neglected compared to the physical response of the system and so can be neglected. Under steady-state conditions, the current $i(t)$ is constant, and equation~(\ref{eq:1}) reduces to :
\begin{equation} \label{eq:3}
\begin{cases}
I_r\dot{\omega}(t)=\tau(t)-\kappa\omega^2(t) \\
v(t)=Ri(t)+\frac{1}{k_v}\omega(t)
\end{cases}	
\end{equation}
where the term $\frac{1}{k_v}\omega(t)$ represent the electromotive force of the motor.
Table~\ref{tb:1}, shows motor and battery coefficients.

A physical Li-ion battery model was given in \cite{mesbahi2013li}. The input of battery model is the current $i_{bat}(t)$, Where the outputs are voltage $V_{bat}(t)$ and state of charge (SoC). This model does not take into account the influence of temperature and the phenomenon of self-discharge. However, it gives results close to reality. The model is based on the two equations, the state of charge (SoC) and The voltage across the cell.
\begin{equation} \label{eq:4}
\begin{cases}
SoC=100\left(1-\frac{\int i_{bat}(t)}{Q_{bat}}\right)\\
V_{bat}=e_m(t)-R_{bat}i_{bat}(t)
\end{cases}	
\end{equation}
$e_m$ is the open circuit voltage (OCV). Its expression is as follow.
\begin{equation} \label{eq:5}
e_m(t)=e_0-k\left(\frac{Q_{bat}}{Q_{bat}-\int i_{bat}(t)}\right)+c_1e^{-c_2\int i_{bat}(t)}
\end{equation}
$e_0$ is the open circuit voltage at full load. $Q_{bat}$ is the cell capacity in Ah, The bias voltage $k$ , exponential voltage $c_1$ and exponential capacity $c_2$ are experimental parameters determined from discharge curve.

\subsection{Quadrotor dynamic model}
The dynamic model for quadrotor aerial vehicle can be derived as follow \cite{Yacef2016}.
\begin{equation} \label{eq:6}
\begin{array}{l}
m\ddot{x} = (\cos\phi \sin\theta \cos\psi + \sin\phi \sin\psi)u_1 \vspace{0.2cm}\\
m\ddot{y} = (\cos\phi \sin\theta \sin\psi - \sin\phi \cos\psi)u_1 \vspace{0.2cm}\\
m\ddot{z} = (\cos\phi \cos\theta)u_1 - mg \vspace{0.2cm}\\
I_x\ddot{\phi} = (I_y-I_z)\dot{\theta}\dot{\psi} + I_r\dot{\theta}\varpi + u_2 \vspace{0.2cm}\\
I_y\ddot{\theta} = (I_z-I_x)\dot{\phi}\dot{\psi} - I_r\dot{\phi}\varpi + u_3 \vspace{0.2cm}\\
I_z\ddot{\psi} = (I_x-I_y)\dot{\phi}\dot{\theta} + u_4	
\end{array}
\end{equation}
where $\varpi=\omega_1-\omega_2+\omega_3-\omega_4$. $I_r$ is the rotor inertia, $m$, $I_x$, $I_y$ and $I_z$ denotes the mass of the quadrotor aerial vehicle and inertia, $l$ is the distance from the center of mass to the rotor shaft, $\omega_j$, $j=1,\ldots,4$ is the rotors velocity, $g=9.81 m/s^2$ is the acceleration due to gravity.

The control inputs are given as follows:
\begin{equation} \label{eq:7}
\begin{array}{l} 
u_1 = \kappa_b(\omega^2_1+\omega^2_2+\omega^2_3+\omega^2_4)\vspace{0.1cm}\\
u_2 = l\kappa_b(\omega^2_2-\omega^2_4)\vspace{0.1cm}\\
u_3 = l\kappa_b(\omega^2_3-\omega^2_1)\vspace{0.1cm}\\
u_4 = \kappa(\omega^2_1-\omega^2_2+\omega^2_3-\omega^2_4)\vspace{0.1cm}
\end{array}
\end{equation}
with $\kappa_b$ is the thrust coefficient. The forces generated by the $i$-th motor is given by $f_i=\kappa_b\omega^2_i$.

\begin{figure}[thpb]
      \centering
      \includegraphics[scale=1]{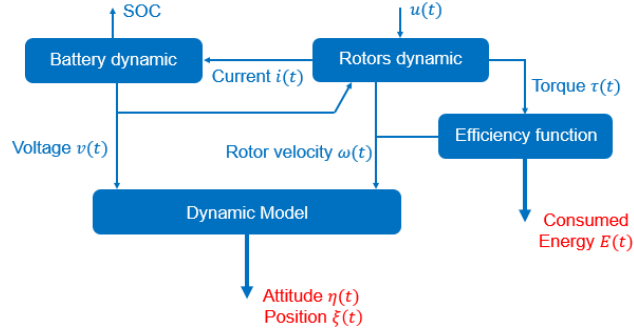}
      \caption{Power lose model for quadrotor aerial vehicle}
      \label{fig.1}
\end{figure}
The amount of energy consumed by the quadrotor during a simple mission between the initial time $t_0$ and the final fixed time $t_f$ is then \cite{yacef2017trajectory}.
\begin{equation}\label{eq:8}
E=\int\limits_{t_0}^{t_f}\sum\limits_{j=1}^4\frac{\Big(I_r\dot{\omega}_j(t)+\kappa\omega^2_j(t)\Big)}{f_{r,j}\big(\dot{\omega}_j(t),\omega_j(t)\big)}\omega_j(t) dt	
\end{equation}
with $f_r(\dot{\omega}(t),\omega(t))$ is efficiency function identified using polynomial interpolation. $\tau_j(t)$ is the torque generated by motor $j$ and $\omega_j(t)$ is rotor velocity at time $t$. Details about efficiency function and parameters identifications can be found in \cite{yacef2017trajectory}.

\begin{figure}[thpb]
      \centering
      \includegraphics[scale=0.5]{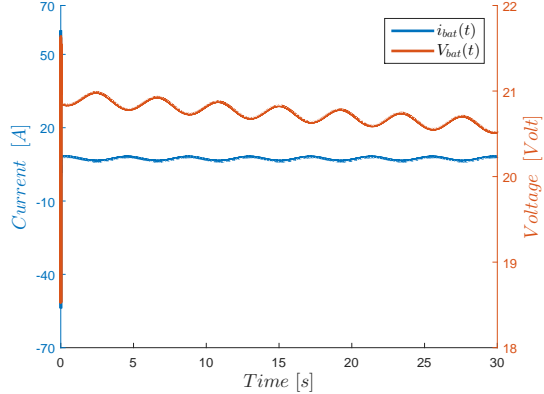}
      \caption{Battery voltage and current }
      \label{fig.2}
\end{figure}

\begin{table}[ht]
\caption{Motor and battery coefficients}
\label{tb:1}
\begin{center}
\begin{tabular}{|c||cl|}
\hline
Parameter&\multicolumn{2}{|c|}{Value}\\
\hline
$I_r$  & $4.1904e^{-5}$&$~kg.m^2$ \\
$k_t$  & $0.0104e^{-3}$&$~N.m/A$ \\
$k_v$  & $96.342$&$~rad/s/volt$ \\
$R$    & $0.2$&$~Ohm$          \\
$Q_{bat}$        & $1.55$&$~Ah$          \\
$R_{bat}$  & $0.02$&$~Ohm$           \\
$e_0$      & $1.24$&$~volt$          \\
$k$        & $2.92e^{-3}$&$~volt$  \\ 
$c_1$        & $0.156$&           \\
$c_2$        & $2.35$&          \\
\hline
\end{tabular}
\end{center}
\end{table} 
\section{Computation of energy-efficiency paths}
In this section we introduce the optimal control problem in order to calculate energy efficiency trajectory for simple mission of quadrotor UAV between a given initial and final configuration. Let $\mathbf{x}=[x_1,\ldots,x_{16}]^T \in\Re^{16}$ denote the state vector, with $x_1=x$,~$x_2=\dot{x}$,~$x_3=y$,~$x_4=\dot{y}$,~$x_5=z$,~$x_6=\dot{z}$,~$x_7=\phi$,~$x_8=\dot{\phi}$,~$x_9=\theta$,~$x_{10}=\dot{\theta}$,~$x_{11}=\psi$,~$x_{12}=\dot{\psi}$,~$x_{13}=\omega_1$,~$x_{14}=\omega_2$,~$x_{15}=\omega_3$,~$x_{16}=\omega_4$ and $\mathbf{u}=[\alpha_1,\alpha_2,\alpha_3,\alpha_4]^T \in\Re^{4}$ the the auxiliary control input vector. We can rewrite system (\ref{eq:6}) in state-space form via dynamic extension,
\begin{equation}\label{eq:9}
\begin{cases}
\dot{x}_1=x_2,~\dot{x}_3=x_4,~\dot{x}_5=x_6\\
\left[
\begin{array}{c}
\dot{x}_2  \\
\dot{x}_4 \\
\dot{x}_6
\end{array} \right]
=-ge_3+\frac{1}{m}\mathbf{F_1}(x_7,x_9,x_{11})\mathbf{B_1}
\left[
\begin{array}{c}
x^2_{13}  \\
\vdots \\
x^2_{16}
\end{array} \right]+\mathbf{W} \\
\dot{x}_7=x_8,~\dot{x}_9=x_{10},~\dot{x}_{11}=x_{12}\\
\left[
\begin{array}{c}
\dot{x}_8  \\
\dot{x}_{10} \\
\dot{x}_{12}
\end{array} \right]
=I_f^{-1}\mathbf{F_2}(x_8,x_{10},x_{12})+I_r\varpi\mathbf{G}
+\mathbf{B_2}\left[
\begin{array}{c}
x^2_{13}  \\
\vdots \\
x^2_{16}
\end{array} \right] \\
\dot{x}_{13}=\alpha_1,~\dot{x}_{14}=\alpha_2,~\dot{x}_{15}=\alpha_3,~\dot{x}_{16}=\alpha_4
\end{cases}
\end{equation}
This yields a nonlinear system affine in the auxiliary control input $\mathbf{u}$. With
$\mathbf{G}=I_f^{-1}[\dot{x}_{10},\dot{x}_8,0]^T$,~$e_3=[0,0,1]^T$,~$\mathbf{B_1}=\kappa_bI_{4\times1}$,~$I_f=diag(I_x,I_y,I_z)$,~$\mathbf{F_1}=[F_{x},F_{y},F_{z}]^T$,~$\mathbf{F_2}=[F_{\phi},F_{\theta},F_{\psi}]^T$ and
$\mathbf{B_2}=\left[
\begin{array}{cccc}
0          &l\kappa_b &0         &-l\kappa_b  \\
-l\kappa_b &0         &l\kappa_b &0 \\
\kappa &-\kappa &\kappa &-\kappa
\end{array} \right]$,$\mathbf{W}=[v_x^{wind},v_y^{wind},0]^T$\\

Now we can cast the energy-efficiency path generation problem as a standard optimal control problem. The final consumed energy $E(tf)$ is used as the cost function. In addition the state vector $\mathbf{x}(t)$ and control input vector $\mathbf{u}(t)$ are constrained to satisfy the vehicle dynamics~(\ref{eq:9}) and boundary conditions. We introduce the following optimal control problem,
$$\min_{\mathbf{u}} E=\int\limits_{t_0}^{t_f}\sum\limits_{j=1}^4\sum\limits_{l=13}^{16}\frac{\Big(I_r\alpha_j(t)+\kappa x^2_l(t)\Big)}{f_{r,j}\big(\alpha_j(t),x_l(t)\big)}x_l(t) dt$$ 
subject to \\
 $~~~~~~~~~~~~~~~~~~~~~~~~~~~~~System (\ref{eq:9})$
\begin{gather}
|x_7|\leq\frac{\pi}{2},~|x_9|\leq\frac{\pi}{2},\nonumber \\
0\leq x_{13}\leq\omega_{max},\ldots,0\leq x_{16}\leq\omega_{max}\nonumber \\
0\leq u_1\leq T_{max},~|u_k|\leq u_{max},~k=1,2,3 \label{eq:10}
\end{gather}
with boundary conditions
\begin{gather}
\mathbf{x}(t_0)= \mathbf{x}_0 \nonumber\\
\mathbf{x}(t_f)= \mathbf{x}_f \nonumber
\end{gather}
where $\omega_{max}$ is the maximum feasible velocity of the aircraft rotors, $T_{max}=4\kappa_b\omega^2_{max}$ is the maximum thrust generated by the quadrotor four rotors and $\mathbf{x}_0,\mathbf{x}_f\in\Re^{16}$. The inequality constraints in (\ref{eq:10}) are associated with physical limitation of vehicle dynamics. 
\section{Numerical experiments}
In order to validate our optimal control approach, we considered the physical parameters of the DJI Phantom 2 quadrotor with multi-rotor propulsion system (2212/920KV motors). The physical parameters of the Phantom 2 used in the simulation experiment, are $l=0.175m$,~$m=1.3kg$,~$I_x=0.081kgm^2$,~$I_y=0.081kgm^2$,~$I_z=0.142kgm^2$,~$\kappa_b=3.8305~10^{-6}N/rad/s$,~$\kappa=2.2518~10^{-8}Nm/rad/s$. We solved the problem (\ref{eq:10}) numerically using the GPOPS-II optimal control software under Matlab 8.5. GPOPS-II software employs  hp-adaptive Gaussian quadrature collocation methods and sparse nonlinear programming \cite{patterson2014}. We used the nonlinear programming (NLP) solver IPOPT (Interior Point OPTimizer) \cite{biegler2003} among the two solvers offered by the software. 

In our first test, we solved problem (\ref{eq:10}) to find the energy-efficiency control inputs $\alpha_i$ allows quadrotor to fly from the origin at time $t_0=0$ to the position $[6,7,8]^T~m$ with the yaw angle take a null value and fixed final time $t_f=10s$. This corresponds to $\mathbf{x}_0=[0_{1\times12},\omega_h,\ldots,\omega_h]^T$ and $\mathbf{x}_f=[6,0,7,0,8,0_{1\times7},\omega_h,\ldots,\omega_h]^T$, where $\omega_h=912 rad/s$ which means that the hovering thrust is $T_h=12,75N$ corresponds the thrust necessary to counterbalance the gravity acceleration. In this case we consider that the quadrotor model (\ref{eq:9}) is not affected by wind gust, which mean that $\mathbf{W}=[0_{1\times3}]^T$. Fig.3 shows the time evolution of the state variables $x_1(t),\ldots,x_{16}(t)$ and control inputs $\alpha_1,\ldots,\alpha_4$ relative to the vehicle path from $\mathbf{x}_0$ to $\mathbf{x}_f$. Fig.4 report the energy-efficiency trajectory of the quadrotor aerial vehicle. Its also reports the trajectory of the quadrotor generated with the adaptive controller considered in \cite{Yacef2016}. Using the optimal control algorithm we obtained a consumption of $1.89~kJ$, where when we used adaptive controller we obtained $5.77~kJ$, which corresponds to a $67.24\%$ increase with respect to the mission trajectory.

In the second test, we consider that system (\ref{eq:9}) is affected by wind gust. As it is well
known, the wind velocity should be modeled in general as a stochastic process \cite{Yacef2018}. In this work, we prefer to use a deterministic wind model so that the same disturbance realization is used in all types of simulations to compare the accuracy of results. 

Thus, the wind velocity along each axis is modeled as a sum of three harmonics plus wind gust according to the expression \cite{mollov2014}
\begin{equation}\label{eq:11}
v^{wind}(t)=v_0+\sum\limits_{k=1}^3A_k sin(\Omega_kt)+v_g(t)
\end{equation}
where
$v_0$ is mean value of wind velocity, $A_k$ amplitude of $k$th harmonic, $\Omega_k$ the frequency of $k$th harmonic and $v_g$ the wind gust.\\
Wind gusts are modeled by the following function
\begin{equation}\label{eq:12}
v_g(t)=\frac{2v_{gmax}}{1+e^{-4(sin(2\pi/T_gt)-1)}}
\end{equation}
where
$v_{gmax}$ is gust amplitude, and $\Omega_g$ gust frequency $(\Omega_g = 2\pi/T_g)$. The wind model parameters used in the simulation are given in Table 2 and the corresponding disturbance actions are shown in Fig. 7.

Now we solved problem (10) to find the energy-efficiency trajectory for quadrotor aerial vehicle under wind conditions. Fig.5 shows the time evolution of the state variables $x_1(t),\ldots,x_{16}(t)$ and control inputs $\alpha_1,\ldots,\alpha_4$ relative to the paths from $\mathbf{x}_0$ to $\mathbf{x}_f$ for this case. Fig.6 report the energy-efficiency trajectory and the quadrotor trajectory generated with the nonlinear adaptive controller (green line) under wind gust conditions. We obtained a consumption of $9.74~kJ$ with the nonlinear adaptive controller, which corresponds to a $78.85\%$ increase with respect to the energy-efficiency trajectory.
\begin{figure*}
      \begin{center}
      \includegraphics[scale=0.38]{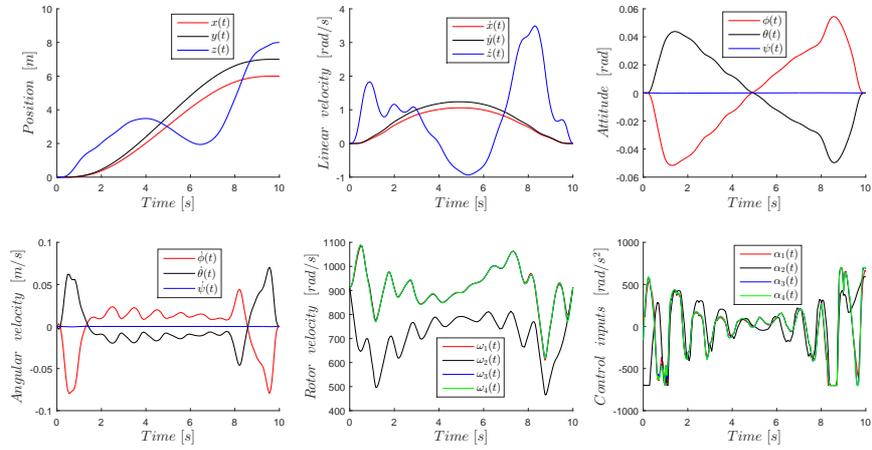}
      \caption{Time evolution of the state variables $x_1(t),\ldots,x_{16}(t)$ and control inputs $\alpha_1,\ldots,\alpha_4$ rotors acceleration, relative to the path from $\mathbf{x}_0$ to $\mathbf{x}_f$.}
      \label{fig.3}
			 \end{center}
\end{figure*}  
\begin{figure}
      \centering
      \includegraphics[scale=0.7]{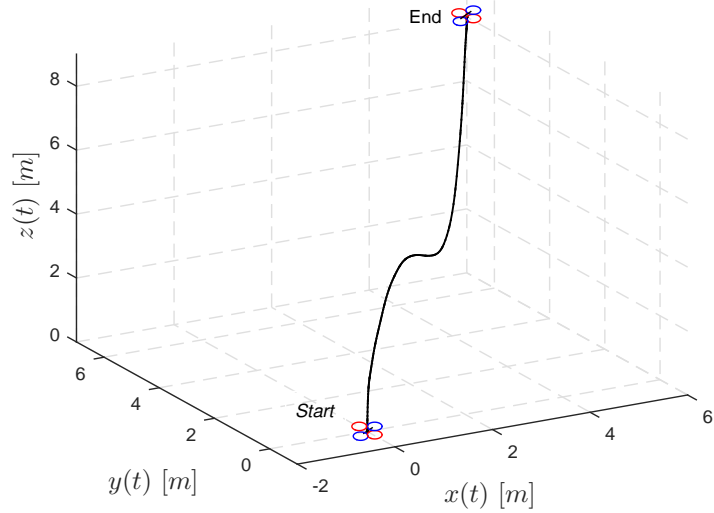}
      \caption{Energy-efficiency 3D trajectory for quadrotor aerial vehicle}
      \label{fig.4}
\end{figure}
\begin{figure*}
      \centering
      \includegraphics[scale=0.38]{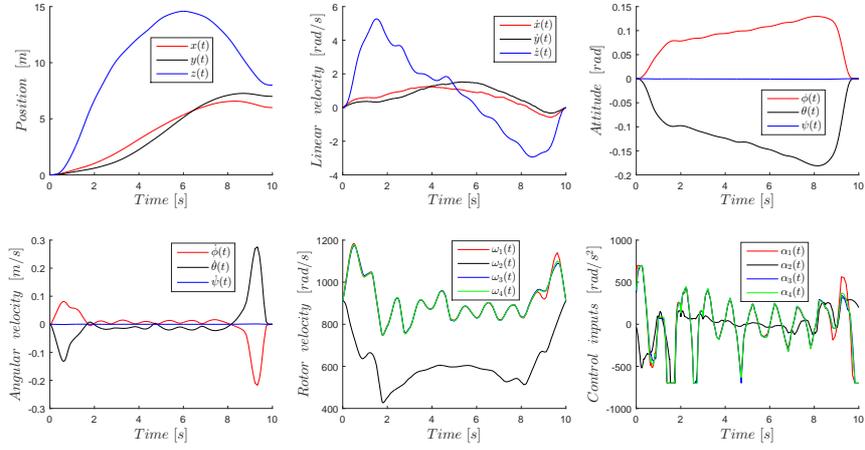}
      \caption{Time evolution of the state variables $x_1(t),\ldots,x_{16}(t)$ and control inputs $\alpha_1,\ldots,\alpha_4$ rotors acceleration, relative to the path from $\mathbf{x}_0$ to $\mathbf{x}_f$ under wind gust conditions}
      \label{fig.5}
\end{figure*}
\begin{figure}
      \centering
      \includegraphics[scale=0.7]{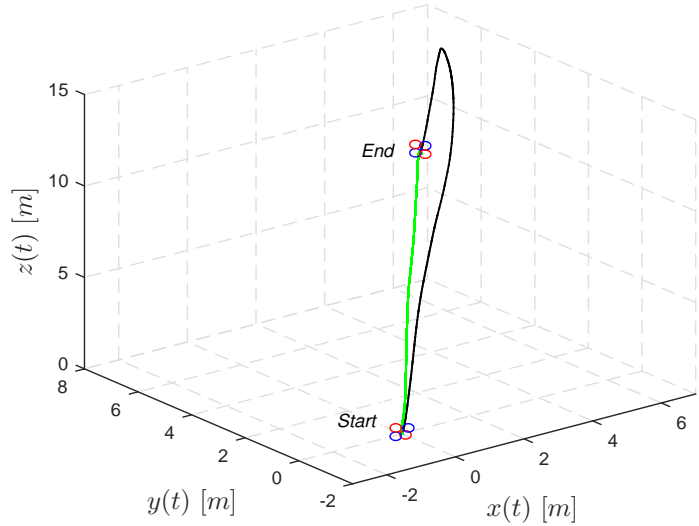}
      \caption{Energy-efficiency 3D trajectory under wind gust conditions}
      \label{fig.6}
\end{figure}
\begin{figure}
      \centering
      \includegraphics[scale=0.5]{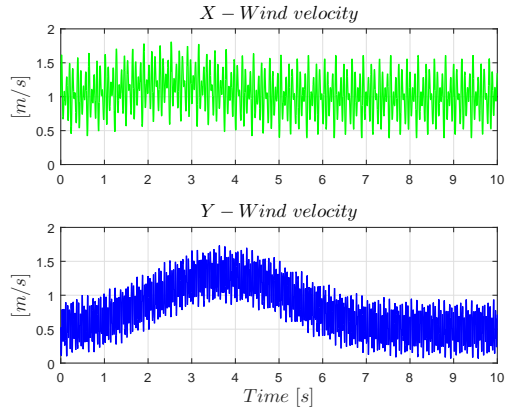}
      \caption{Wind Velocity}
      \label{fig.7}
\end{figure}
\begin{table}[ht]
\caption{Wind model parameters}
\label{tb:2}
\begin{center}
\begin{tabular}{|c||c||c||c||c||c||c|}
\hline
Axis&$v_0$   &k   &$\Omega_k$ &$A_k$   &$v_{gmax}$ &$T_g$\\
    &$(m/s)$ &    &$(Hz)$     &$(m/s)$ &$(m/s)$    &$(s)$\\
\hline
$X$ &$1.0$   &$1$ &$0.5$      &$0.10$  &$0.20$     &$10$ \\
    &        &$2$ &$0.7$      &$0.25$  &           &     \\
    &        &$3$ &$1.0$      &$0.30$  &           &     \\
$Y$ &$0.5$   &$1$ &$0.6$      &$-0.05$  &$0.20$     &$10$ \\
    &        &$2$ &$1.0$      &$-0.10$  &           &     \\
    &        &$3$ &$1.5$      &$-0.30$  &           &     \\        
\hline
\end{tabular}
\end{center}
\end{table}  
\section{Conclusion}
The central aim of this study was to determine the optimal trajectory and control inputs in terms of energy consumption for a quadrotor aerial vehicle to travel from an initial hover configuration to final hover configuration under windy conditions. In a first step we considered a power lose model to compute the consumed mechanical energy, then we have calculated energy-efficiency trajectories between two given boundary states, by solving an optimal control problem. We have compared The energy consumed by the vehicle with the optimal control algorithm with an adaptive control approach in order to evaluated the mount of energy saved for quadrotor aerial vehicle.

In future works, we plan to introduce aerodynamic effects in the power lose model, and use an on-board MPC controller to obtained minimum energy consumption for quadrotor aerial vehicle mission.     
%
%
%
\bibliographystyle{IEEEtran}
\bibliography{refbibliography}

\end{document}